\documentclass{article}

\usepackage{arxiv}

\usepackage[utf8]{inputenc} 
\usepackage[T1]{fontenc}    
\usepackage{hyperref}       
\usepackage{url}            
\usepackage{booktabs}       
\usepackage{amsfonts}       
\usepackage{nicefrac}       
\usepackage{microtype}      
\usepackage{lipsum}
\usepackage{graphicx}
\usepackage{multirow}
\graphicspath{ {./fig/} }

\usepackage{biblatex}
\title{On Measuring Semantic Preservation in Legal Ontology Learning}

\author{
 Albert Sadowski \\
  Warsaw University of Technology \\
  Warsaw, Poland \\
  \texttt{albert.sadowski.stud@pw.edu.pl} \\
   \And
 Jarosław A. Chudziak \\
  Warsaw University of Technology \\
  Warsaw, Poland \\
  \texttt{jaroslaw.chudziak@pw.edu.pl} \\
}

\begin{document}
\maketitle
\begin{abstract}
Ontology learning transforms unstructured text into structured representations for automated reasoning. Yet structuring information risks losing it, and current evaluation methodologies cannot detect such loss, focusing on structural correctness while failing to measure whether meaning survives transformation. We propose an evaluation methodology that addresses this: comparing LLM task performance on source documents against performance on transformed representations, with the difference quantifying semantic loss. We demonstrate this approach on legal merger agreement analysis, a domain chosen for its complex language and precise semantic requirements, comparing direct LLM application against three ontology learning methods across six language models. The results reveal systematic semantic loss with significant variation based on reasoning complexity and model-method interactions. Our contributions are: (1) an evaluation framework for measuring semantic preservation in ontology learning, and (2) empirical evidence that semantic loss varies dramatically with model-method pairing, providing guidance for selecting optimal configurations in legal knowledge systems.
\end{abstract}


\section{Introduction}

Ontology learning seeks to automatically transform unstructured text into structured, machine-readable representations that can support reasoning and integration across systems \cite{Buitelaar2005OntologyLF, maedche2002ontology}. This approach has attracted renewed interest in specialized domains, particularly in law \cite{RODRIGUES201912,casellas2011legal}, where the volume and complexity of textual information creates challenges for manual knowledge organization efforts.

Ontologies play a complementary role alongside large language models (LLMs) in modern knowledge systems. While natural language serves humans well, it functions poorly as a communication protocol between computational systems due to inherent ambiguity and inconsistency \cite{ontologiesIntheEraOfLLMs}. Ontologies provide controlled vocabularies essential for data interoperability, transparent knowledge representations, and logical consistency requirements that LLMs cannot guarantee \cite{ontologiesIntheEraOfLLMs}. This creates opportunities for hybrid architectures where LLMs interface with symbolic reasoning systems through structured ontological representations \cite{sadowskiChudziak, kostkaChudziak}.

However, a critical question remains underexplored: when we transform natural language text into ontological representations, do we lose the semantic content needed for downstream tasks? Current evaluation methodologies focus on structural correctness \cite{brank2005survey,brewster-etal-2004-data}, assessing whether ontologies conform to formal specifications and exhibit logical consistency. These measures ensure technical validity but cannot detect semantic loss. An ontology may pass all structural tests yet fail to support the reasoning tasks for which it was intended. Without methods to measure semantic preservation, practitioners cannot make evidence-based decisions about when ontological transformation helps or hinders their applications.

\begin{figure*}[h!]
  \centering
  \includegraphics[width=0.9\textwidth]{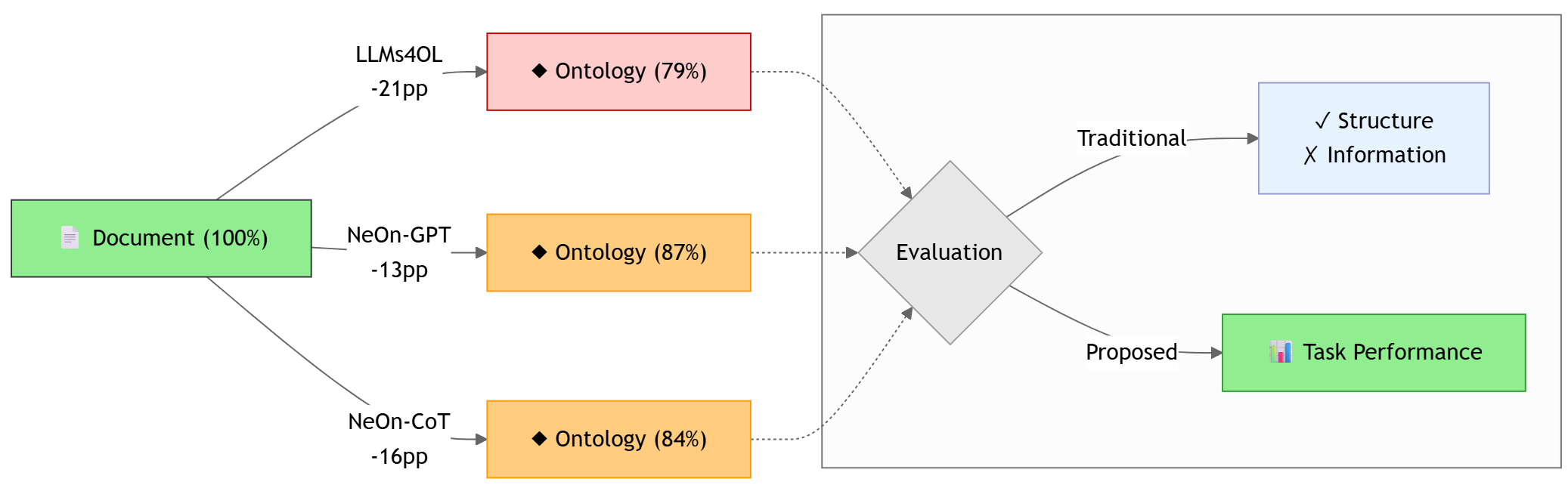}
  \caption{Overview of our evaluation methodology. Left: source documents (baseline 100\% accessible content) are transformed via three ontology learning methods, each introducing semantic loss (average accuracy, LLMs4OL: -21pp, NeOn-GPT: -13pp, NeOn-CoT: -16pp). Right: traditional evaluation validates structure but cannot detect information loss; our proposed task-based approach measures semantic preservation by comparing LLM accuracy on identical tasks before and after transformation.}
  \label{fig:funnel}
\end{figure*}

We propose an evaluation methodology that addresses this gap by using LLM performance as a proxy for semantic accessibility. Our approach treats baseline LLM performance on source documents as a model-specific upper bound on accessible semantic content, then measures how much of that content remains accessible after ontological transformation. This methodology builds on task-based evaluation paradigms from NLP \cite{sparckGalliers1996,porzelMalaka2004} and recent behavioral testing frameworks \cite{ribeiro-etal-2020-beyond}. Rather than claiming to measure absolute semantic content, we quantify the relative preservation of information that a given model can extract, enabling direct comparison between source documents and their ontological transformations under consistent conditions.

We demonstrate this approach using the MAUD dataset from LegalBench \cite{legalBench,maud}, comparing direct LLM application against three ontology learning methods: LLMs4OL \cite{babaei2023llms4ol}, NeOn-GPT \cite{fathallah2024neon}, and a simplified NeOn-CoT method we introduce. Our evaluation spans six language models to ensure robust findings across different architectures. Results reveal systematic semantic loss across all approaches, with particularly severe degradation for complex legal reasoning tasks.

The contributions of this work are: (1) an evaluation framework for measuring semantic preservation in ontology learning, and (2) empirical evidence that semantic loss in legal ontology learning varies dramatically with model-method pairing, providing guidance for selecting optimal configurations.

The remainder of this paper is organized as follows. Section~\ref{sec:related-work} reviews related work. Section~\ref{sec:methodology} presents the proposed methodology, the three ontology learning approaches, and the experimental setup. Section~\ref{sec:results} reports results, including a task-level analysis. Sections~\ref{sec:discussion} and~\ref{sec:limitations} discuss implications, limitations, and future work. Section~\ref{sec:summary} concludes.

\section{Related Work}
\label{sec:related-work}

Ontology engineering has evolved from traditional manual approaches \cite{casellas2011legal, RODRIGUES201912} to recent LLM-based automated methods \cite{babaei2023llms4ol, fathallah2024neon}. Traditional approaches used lexico-syntactic pattern mining and clustering \cite{hearst1998automated}, frameworks like Text2Onto \cite{cimiano2005text2onto}, and the NeOn methodology for structured ontology development \cite{suarez2015neon}. Early work explored semantic nets as meta-level representations for maintaining semantic accessibility across heterogeneous data types \cite{chudziakPiotrowski}. Legal ontology projects drew on jurisprudential foundations and employed bottom-up, top-down, and hybrid development strategies \cite{RODRIGUES201912}. Construction methodologies included Methontology \cite{corcho2005building}, Ontology Development 101 \cite{noy2001ontology}, and CommonKADS \cite{valente1999legal}. Systematic analysis showed 93.5\% of legal ontology development was manual, with 41\% lacking validation activity \cite{RODRIGUES201912}. Evaluation followed Gold Standard, Case-Based, Data-Driven, and User-Based approaches \cite{brank2005survey}. The legal domain posed challenges including complex writing styles, heterogeneous sources, and evolving documents that contributed to syntactic and semantic anomalies \cite{RODRIGUES201912}.

LLMs introduced new paradigms based on automated knowledge extraction \cite{babaei2023llms4ol}. LLMs4OL addresses term typing (MAP@1), taxonomy discovery, and relation extraction (F1-scores), reporting term typing as the most challenging task (60-80\% baseline) with fine-tuning gains of 25\%, 18\%, and 3\% respectively \cite{babaei2023llms4ol}. NeOn-GPT combines structured methodology with LLM capabilities, using structural metrics such as axiom counts and hierarchical depth \cite{fathallah2024neon}. Both approaches focus on structural correctness rather than knowledge representation quality or semantic preservation \cite{babaei2023llms4ol,fathallah2024neon}. Structural metrics fail to capture semantic richness or semantic loss during transformation, particularly in specialized domains like law \cite{RODRIGUES201912}.

Task-based evaluation offers an alternative perspective.\label{section:task-based} Sparck Jones and Galliers established the distinction between intrinsic and extrinsic evaluation \cite{sparckGalliers1996}, and Mollá and Hutchinson showed disconnects between structural accuracy and downstream performance \cite{mollaHutchinson2023}. For knowledge systems, Porzel and Malaka pioneered task-based ontology evaluation \cite{porzelMalaka2004}, extended by Gangemi et al. \cite{gangemi2025} and Brewster et al. \cite{brewster-etal-2004-data}. Similar to behavioral testing frameworks like CheckList \cite{ribeiro-etal-2020-beyond}, we treat LLM performance as a signal for evaluating representation quality.

Information-theoretic approaches provide quantitative frameworks for measuring preservation during representation transformations. Knowledge distillation research introduced loyalty measures using Jensen-Shannon divergence \cite{xu2021beyond} and methods for measuring semantic retention across representational levels \cite{jiao2020tinybert}. Multi-scale mutual information methods show that preservation must be measured across granularities to capture local and global coherence \cite{zhao2019region}. LLM-as-evaluator studies show strong LLMs reach over 80\% agreement with human preferences \cite{zheng2023judging}, though bias studies note limitations that require careful calibration \cite{wang2023evaluators}. Ensemble approaches with diverse LLM evaluators show promise for reliable assessment \cite{verga2024replacing}. These results support our use of baseline LLM performance as a reference point for measuring semantic preservation during ontological transformation.

\section{Methodology}
\label{sec:methodology}

Current ontology learning evaluation methods focus on structural correctness rather than knowledge transformation quality \cite{brank2005survey, babaei2023llms4ol, fathallah2024neon}. Building on the task-based evaluation paradigm from NLP (Section~\ref{section:task-based}), we propose an evaluation framework that quantifies semantic preservation by using LLM performance on downstream tasks as a proxy for semantic accessibility \cite{porzelMalaka2004, ribeiro-etal-2020-beyond}.

Our approach treats baseline LLM performance on source documents as the total accessible content extractable by that LLM, defined as $LLM(text, question) \rightarrow answer$ performance on original legal text. This baseline establishes an upper bound for semantic accessibility from that model's perspective, following the principle that extrinsic task performance provides a more meaningful signal than intrinsic structural measures \cite{sparckGalliers1996, mollaHutchinson2023}.

While ontological transformation may intentionally prioritize structural organization and formal reasoning capabilities over complete semantic preservation, measuring accessibility trade-offs enables informed decisions about when ontological transformation enhances versus diminishes practical utility for specific applications. We quantify this as: $SemanticLoss = BaselinePerf - TransformedPerf$.

This framework provides domain-agnostic assessment applicable across legal specializations without requiring domain-specific gold standards, measures semantic preservation from a user-centric perspective focusing on practical utility for downstream applications, and enables systematic comparison of multiple ontology learning approaches. Importantly, our methodology measures semantic preservation relative to each LLM's capabilities, accounting for model-specific baseline variations.

\subsection{Ontology Learning Approaches}

We evaluate three ontology learning approaches against the direct LLM application baseline established in the previous section. Each approach transforms source legal documents into structured ontological representations using different computational strategies, representing different levels of methodological structure and computational complexity, as well as varying degrees of domain awareness. The approaches vary in complexity from parallel task decomposition to sequential methodology-guided construction, and differ in their degree of domain targeting - from completely domain-agnostic (LLMs4OL) to explicitly domain-aware (NeOn-GPT and NeOn-CoT). This variation enables systematic analysis of how both structural complexity and domain specificity affect semantic preservation during ontological transformation.

\paragraph{LLMs4OL} Based on the framework proposed by Babaei Giglou et al. \cite{babaei2023llms4ol}, this approach decomposes ontology construction into three parallel tasks: term typing (classifying significant terms), taxonomy discovery (identifying hierarchical relationships), and non-taxonomic relation extraction (discovering semantic relationships between entities). The approach is designed to be domain-agnostic, making no assumptions about the specific domain or downstream evaluation tasks. Since the original LLMs4OL framework focuses on individual ontology learning tasks without integration, we extend the approach with an additional integration step that synthesizes results from all three tasks into a complete OWL ontology in Turtle format. This integration step uses zero-shot prompting to combine term classifications, hierarchical relationships, and semantic relationships into a coherent ontological representation without domain-specific guidance.

\paragraph{NeOn-GPT} Following the methodology described by Fathallah et al. \cite{fathallah2024neon}, this approach implements a structured five-stage pipeline: requirements specification, competency question generation, conceptual model creation, ontology draft implementation, and enrichment with instances and annotations. Unlike LLMs4OL, this approach explicitly incorporates domain awareness by identifying the text as mergers and acquisitions (M\&A) contract fragments and steering the ontology construction process toward legal document representation. While the original NeOn-GPT includes validation mechanisms (syntax checking, consistency verification, and pitfall resolution), we implement the core methodology stages without the validation loops, as syntactic correctness is less critical for our semantic preservation evaluation than substantive knowledge representation.

\paragraph{NeOn-CoT} This approach consolidates the NeOn methodology principles into a single chain-of-thought prompting strategy. Similar to NeOn-GPT, it maintains domain awareness by explicitly referencing the legal context and M\&A domain. The prompt guides the LLM through the conceptual phases of ontology development (domain analysis, competency questions, entity extraction, conceptual modeling, and formal implementation) within a unified processing step, reducing computational overhead while maintaining the structured reasoning approach.

\subsection{Representation Challenges and Constraints}

Our approach adopts two methodological constraints. First, LLMs process ontological representations as textual strings rather than formal logical structures. This aligns with our goal of measuring semantic accessibility from a user-centric perspective, reflecting how practitioners typically interact with these representations.

Second, the framework operates under information conservation principles: ontology learning approaches reorganize existing content without external knowledge augmentation. This enables attribution of performance differences to transformation effects rather than knowledge enrichment. Suboptimal baseline LLM performance reflects model-specific interpretation capabilities rather than insufficient source information.

These choices complement rather than replace evaluations of ontologies' formal computational capabilities.

\subsection{Dataset and Experimental Setup}

Our evaluation employs the Merger Agreement Understanding Dataset (MAUD) from LegalBench \cite{legalBench,maud}. MAUD contains 34 multiple-choice tasks focusing on merger agreement analysis, including material adverse effect definitions, representations and warranties, and fiduciary obligations (see Appendix~\ref{sec:maud_tasks} for complete task listing). We use all 34 tasks. To ensure balanced evaluation, we selected 69 examples from each task (the minimum available across all tasks), yielding 2,346 total examples.

We evaluate across six language models: o4 mini (\href{https://platform.openai.com/docs/models/o4-mini}{\textit{o4-mini-2025-04-16}}), GPT 4.1 mini (\href{https://platform.openai.com/docs/models/gpt-4.1-mini}{\textit{gpt-4.1-mini-2025-04-14}}), Claude Sonnet 4 (\href{https://docs.anthropic.com/en/docs/about-claude/models/overview}{\textit{claude-sonnet-4-20250514}}), Gemini 2.5 Flash (\href{https://ai.google.dev/gemini-api/docs/models#gemini-2.5-flash}{\textit{gemini-2.5-flash-preview-05-20}}), \href{https://fireworks.ai/models/fireworks/deepseek-v3}{\textit{DeepSeek v3}}, and \href{https://fireworks.ai/models/fireworks/llama4-maverick-instruct-basic}{\textit{Llama 4 Maverick}}. Testing across multiple models allows us to assess whether semantic loss patterns generalize across architectures.

For each model, we measure accuracy on the multiple-choice tasks under two conditions: (1) baseline, where the model answers directly from source text using zero-shot prompting, and (2) ontology-transformed, where the model first generates an ontology, then answers from that ontology. Each model generates its own ontologies, ensuring consistent capabilities between transformation and evaluation. Temperature is set to 0 where supported. Answers are extracted via structured outputs. We report single-run results; statistical significance is established across the 2,346 test examples.

The implementation uses Python 3.13 with \href{https://github.com/langchain-ai/langchain}{LangChain 0.3.25} and \href{https://www.langchain.com/langgraph}{LangGraph 0.4.8}. Models were accessed through the respective provider APIs: OpenAI, Anthropic, Google Gemini API, and Fireworks AI (DeepSeek v3 and Llama 4 Maverick). Complete prompts and the code are available in an online code repository\footnotemark[1].

\footnotetext[1]{\url{https://github.com/albsadowski/ontology-learning-eval}}

\section{Results}
\label{sec:results}

Our experimental results reveal substantial semantic loss across all ontology learning approaches, with performance degradation ranging from 8.4 percentage points (pp) to 27.4pp compared to baseline accessibility levels. Table~\ref{tab:information_loss} presents semantic loss measurements calculated as the percentage reduction from baseline performance for each model-method combination.

\begin{table*}[h]
\caption{Information Preservation Analysis Across Models and Ontology Learning Methods}
\label{tab:information_loss}
\footnotesize
\begin{tabular*}{\hsize}{@{\extracolsep{\fill}}l l c c c@{}}
\toprule
\textbf{Model} & \textbf{Method} & \textbf{Loss (pp)} & \textbf{Success (\%)} & \textbf{Failure (\%)} \\
\midrule
\multirow{3}{*}{GPT 4.1 mini (64.51\%)} & LLMs4OL & 17.5 & 64.13 & 16.02 \\
 & \textbf{NeOn-GPT} & \underline{\textbf{8.8}} & \underline{\textbf{79.33}} & 12.71 \\
 & NeOn-CoT & 12.7 & 71.73 & 15.47 \\
\addlinespace[0.3em]
\multirow{3}{*}{Gemini 2.5 Flash (75.14\%)} & LLMs4OL & 27.4 & 57.71 & 16.97 \\
 & NeOn-GPT & 11.2 & 81.39 & 8.98 \\
 & \textbf{NeOn-CoT} & \underline{\textbf{8.4}} & \underline{\textbf{82.9}} & 16.02 \\
\addlinespace[0.3em]
\multirow{3}{*}{DeepSeek V3 (69.53\%)} & LLMs4OL & 25.1 & 54.31 & 22.01 \\
 & \textbf{NeOn-GPT} & \textbf{17.6} & \textbf{65.82} & 20.08 \\
 & NeOn-CoT & 25.3 & 53.3 & 23.55 \\
\addlinespace[0.3em]
\multirow{3}{*}{Claude Sonnet 4 (59.80\%)} & LLMs4OL & 24.8 & 45.08 & 19.51 \\
 & \textbf{NeOn-GPT} & \textbf{18.1} & \textbf{60.66} & 13.41 \\
 & NeOn-CoT & 19.1 & 53.28 & 21.95 \\
\addlinespace[0.3em]
\multirow{3}{*}{o4 mini (70.78\%)} & LLMs4OL & 22.8 & 61.94 & 14.09 \\
 & \textbf{NeOn-GPT} & \underline{\textbf{9.6}} & \underline{\textbf{80.06}} & 15.44 \\
 & NeOn-CoT & 16.3 & 70.9 & 14.77 \\
\addlinespace[0.3em]
\multirow{3}{*}{Llama 4 Maverick (65.53\%)} & \textbf{LLMs4OL} & \textbf{13.9} & 59.29 & 36.22 \\
 & \textbf{NeOn-GPT} & 14.5 & \textbf{64.8} & 24.19 \\
 & NeOn-CoT & 19 & 57.17 & 25.05 \\
\bottomrule
\end{tabular*}
\vspace{0.3em}

\footnotesize
\textit{Note:} Loss = information loss (pp). Success = success case accuracy. Failure = failure case accuracy. Bold = best per model; underlined = overall top performers.
\end{table*}

LLMs4OL exhibits the most severe semantic loss (13.9-27.4pp), likely due to information fragmentation during parallel task decomposition. NeOn-GPT demonstrates the best preservation characteristics (8.8-18.1pp.), suggesting that explicit domain awareness and sequential processing better preserve legal document semantics. NeOn-CoT shows intermediate performance with notable variability (8.4-25.3pp.), trading some preservation benefits for computational efficiency.

When baseline models successfully extract information, ontology learning methods retain 45-83\% of accessible content, with NeOn-GPT achieving the highest success preservation rates (60-83\%). Analysis of failure cases reveals limited compensatory benefits from ontological transformation, with most approaches showing only 9-36\% accuracy on baseline failures, indicating that ontology learning primarily reorganizes existing accessible content rather than making previously inaccessible information available.

\paragraph{Cross-Model Consistency}

The semantic loss patterns demonstrate notable consistency across diverse model architectures, providing evidence that observed degradation reflects systematic limitations of current ontology learning approaches rather than model-specific artifacts.

The performance hierarchy NeOn-GPT $>$ NeOn-CoT $>$ LLMs4OL holds across all models tested, despite architectural differences between transformer variants (GPT 4.1 mini, Claude Sonnet 4), mixture-of-experts systems (DeepSeek v3), and reasoning-enhanced models (o4 mini). This consistency suggests structured, domain-aware approaches systematically outperform parallel task decomposition methods for legal ontology learning.

However, absolute sensitivity varies considerably across models. Claude Sonnet 4 shows higher sensitivity (45-61\% preservation rates), while Gemini 2.5 Flash demonstrates more robust preservation (57-83\% rates), suggesting model selection represents a critical factor for practical applications. Despite baseline performance differences (59.8\% to 75.14\%), all models exhibit comparable semantic loss magnitudes, with standard deviation below 0.09 for each method, indicating the transformation process imposes consistent accessibility constraints regardless of initial model capabilities.

All performance differences between baseline and ontology learning methods were statistically significant (Wilcoxon signed-rank tests, $p < 0.001$), with effect sizes ranging from medium ($d = 0.53$ for NeOn-GPT) to large ($d = 0.85$ for LLMs4OL). Pairwise comparisons confirmed the observed performance hierarchy, with all method differences significant ($p < 0.001$).

\paragraph{Semantic Loss Analysis}
\label{section:analysis}

\begin{figure}[t]
  \centering
  \includegraphics[width=0.65\columnwidth]{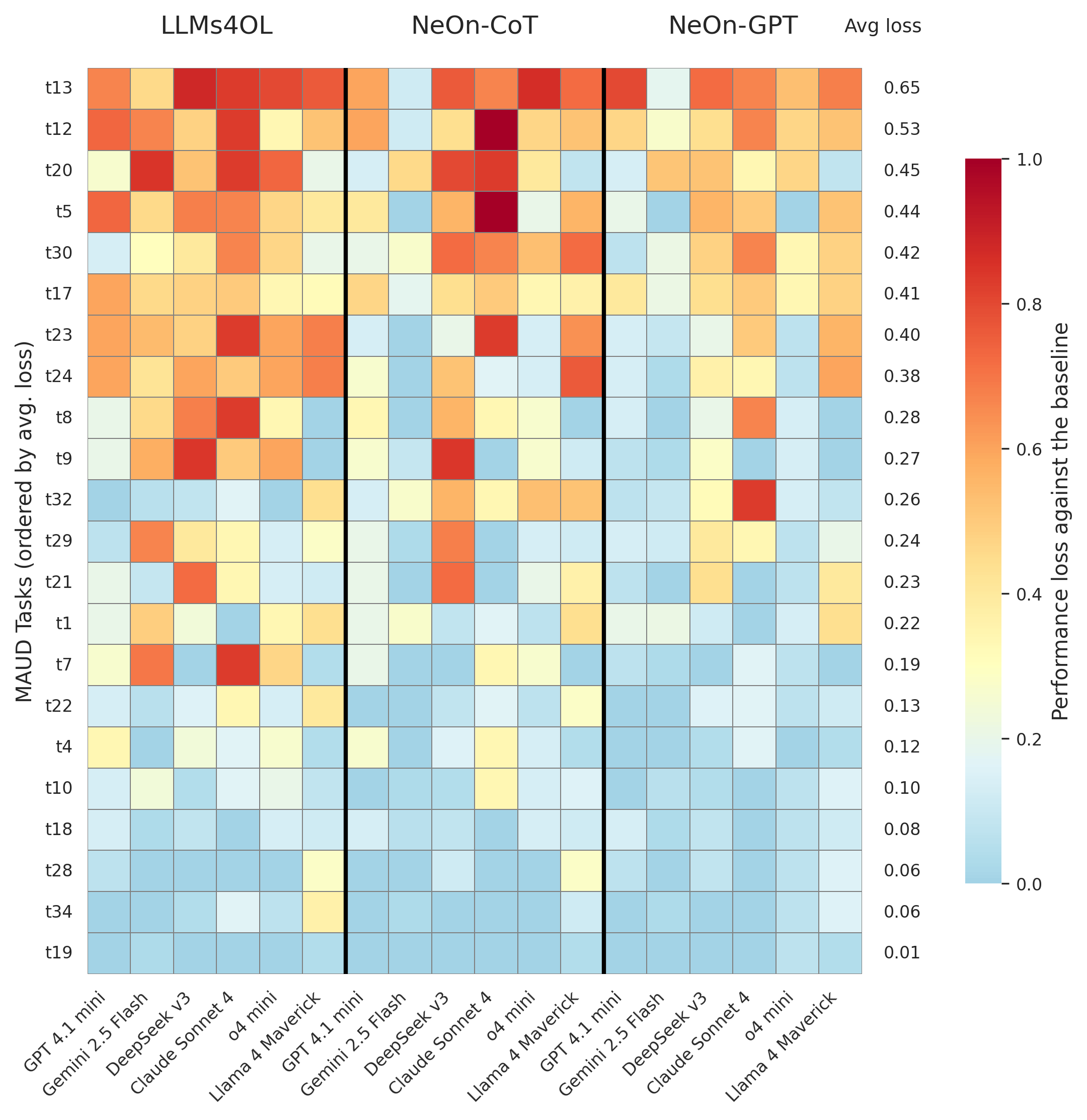}
  \caption{Performance loss across the subset of MAUD tasks with baseline accuracy $\geq 50\%$; tasks below this threshold (e.g., t11) are excluded from the heatmap but remain part of the aggregate results in Table~\ref{tab:information_loss}. The full task list is given in~Appendix~\ref{sec:maud_tasks}.}
  \label{fig:heatmap}
\end{figure}

To explore patterns of semantic loss in legal ontology learning, we analyzed MAUD tasks with sufficient baseline performance ($\geq 50\%$) to ensure meaningful statistical analysis. Figure~\ref{fig:heatmap} reveals systematic patterns varying by linguistic and logical complexity, ranging from minimal degradation for straightforward categorical determinations to severe semantic loss (up to 65\% average loss) for complex multi-standard legal reasoning tasks. Critically, significant model-method interactions emerge, with Gemini demonstrating exceptional performance when paired with NeOn methods, particularly NeOn-CoT, suggesting optimal ontology learning requires careful model-method matching.

The most severe semantic loss occurs with nuanced legal standards and modal language defining procedural requirements. Task \textit{t13} (fiduciary duty standards for superior offer recommendations) shows the highest average loss at 65\%, followed by \textit{t12} (fiduciary duty standards for intervening event recommendations) at 53\%. This illustrates a challenge: while source documents specify that boards may act when continuation would more likely than not result in a violation of fiduciary duties, ontological representations abstract this to generic concepts like $LikelyViolationOfFiduciaryDuties$, losing precise probabilistic language (more likely than not vs.\ reasonably likely vs.\ could reasonably be expected) that distinguishes legal standards.

However, model-method interactions reveal striking differences. While \textit{t13} shows devastating loss with most combinations, Gemini paired with NeOn-CoT achieves 88\% semantic retention. Similarly, \textit{t12} shows 53\% average loss, but Gemini with NeOn-CoT maintains 86\% baseline performance compared to just 13\% with LLMs4OL, illustrating how modal expressions respond dramatically differently to various ontological approaches.

Conversely, tasks involving straightforward binary determinations demonstrate near-perfect preservation across all combinations. Task \textit{t19} (whether financial point of view is sole consideration for superior offers) achieved best preservation with only 1\% average loss, followed by \textit{t34} (consideration type classification, 6\% loss) and \textit{t28} (efforts standard determination, 6\% loss). These leverage ontologies' strength in capturing explicit relationships while avoiding linguistic nuance causing degradation in complex reasoning tasks.

The pronounced model-method interaction effects reveal that effectiveness depends critically on appropriate pairing. Tasks involving complex multi-standard determinations show 45-65\% average losses, but Gemini consistently outperforms other models with NeOn methods, winning 9 out of 10 challenging tasks analyzed. This suggests Gemini's architecture is particularly suited to NeOn's structured reasoning patterns, potentially due to superior handling of chain-of-thought and step-by-step ontological reasoning.

Method-specific analysis reveals the sophistication hierarchy (NeOn-GPT $>$ NeOn-CoT $>$ LLMs4OL) applies primarily to certain model types, with Gemini representing a clear outlier. While NeOn-GPT generally achieves best average preservation across all models, Gemini specifically excels with NeOn-CoT, suggesting synergistic effects for legal ontology learning. This indicates future research should focus on identifying optimal model-method pairings rather than pursuing universal approaches, with Gemini + NeOn-CoT representing a particularly promising direction for maintaining semantic fidelity in legal ontology transformations.

\section{Discussion}
\label{sec:discussion}

The consistency of semantic loss across diverse models and methods suggests this may reflect characteristics of current transformation approaches rather than implementation-specific artifacts.

These findings suggest the need for more evidence-based approaches to ontological transformation, moving from assumptions about structural benefits to empirical validation of when ontologies are appropriate for specific applications. Semantic loss appears intrinsic to the abstraction process - when ontological representations transform ``more likely than not'' into generic concepts like $LikelyViolationOfFiduciaryDuties$, critical legal distinctions vanish irreversibly. This is not a technical problem to be solved through better algorithms, but a fundamental characteristic of structural abstraction that practitioners must acknowledge.

The discovery that optimal performance requires specific model-method pairings (e.g., Gemini + NeOn-CoT achieving 86-88\% semantic retention while other combinations preserve only 12-45\%) suggests ontology learning is not universally applicable. Rather than pursuing one-size-fits-all approaches, the field should focus on identifying optimal configurations for specific legal reasoning tasks and content types.

Our LLM-based evaluation methodology could represent a useful complement to existing ontology assessment approaches. While structural correctness metrics ensure technical validity, they may not fully capture the semantic richness essential for downstream applications. Semantic preservation metrics could supplement structural measures, enabling more evidence-based decisions about when ontological transformation enhances versus diminishes practical utility.

Despite systematic semantic loss during transformation, ontologies remain important in the LLM era for bridging neural and symbolic systems. Natural language functions poorly as a communication protocol between computational systems due to inherent ambiguity and inconsistency \cite{ontologiesIntheEraOfLLMs}. Ontologies provide controlled vocabularies essential for data interoperability, transparent knowledge representations, and logical consistency requirements that LLMs cannot guarantee \cite{ontologiesIntheEraOfLLMs}. This complementary relationship suggests ontologies' role as intermediaries in hybrid architectures, enabling LLMs to interface with symbolic reasoning systems, databases, and automated decision-making processes that require formal guarantees.

\section{Limitations and Future Work}
\label{sec:limitations}

Several limitations should temper the interpretation of these results. The evaluation covers a single legal subdomain, merger and acquisition contract analysis through MAUD, so the magnitude of semantic loss and the observed NeOn-GPT $>$ NeOn-CoT $>$ LLMs4OL hierarchy may not transfer to other legal subdomains or to non-legal domains. The task format is multiple-choice, which is convenient for automated scoring but bounds the semantic distinctions the framework can surface; open-ended legal tasks may exhibit different patterns. Coverage of ontology learning approaches is limited to three methods.

The framework itself relies on baseline LLM performance as a proxy for accessible content, so semantic loss is measured relative to each model's capabilities rather than against ground-truth semantic content; results should be read as comparative rather than absolute. Ontologies are consumed by LLMs as text, which means the framework does not credit benefits that would materialize only through formal reasoning such as SPARQL queries or automated inference. We also assume information conservation, that is, ontology generation introduces no external knowledge; approaches that intentionally enrich ontologies fall outside the framework as defined here.

Given the systematic semantic loss observed across all transformation approaches, future research should investigate how ontological representations might better preserve complex legal standards. Tasks like fiduciary duty determinations showed the highest degradation, suggesting that current methods struggle with the precise formulations that distinguish legal standards in practice.

Domain generalization studies should test whether observed patterns hold across legal specializations and other knowledge domains where precise language matters. Computational context evaluation should also assess whether semantic loss is offset by formal reasoning capabilities through SPARQL queries and automated inference within ontologies' intended computational environment \cite{brank2005survey}.

Selective transformation approaches could optimize the accessibility-structure trade-off by applying ontological methods only to content types that benefit from formal representation while preserving linguistically nuanced passages in natural language. Finally, systematic model-method optimization should explore architecture-methodology interactions to identify optimal pairings for specific legal reasoning requirements \cite{colelough2025neurosymbolicai2024systematic}.

\section{Summary}
\label{sec:summary}

Ontology learning transforms complex documents into structured representations for automated reasoning. Yet structuring information risks losing it, and current evaluation methodologies cannot detect such loss, focusing on structural correctness while failing to measure whether meaning survives transformation. We propose an evaluation methodology that addresses this gap: comparing LLM task performance on source documents against performance on transformed representations, with the difference quantifying semantic loss. This approach treats baseline performance as accessible content and enables evidence-based assessment of when ontological transformation helps or hinders downstream applications.

Evaluating this methodology on legal merger agreement analysis using the MAUD dataset, we compare direct LLM application against three ontology learning approaches (LLMs4OL, NeOn-GPT, and NeOn-CoT) across six state-of-the-art language models. Our results reveal systematic semantic loss across all approaches, with severe losses for complex legal reasoning tasks involving nuanced modal language, while simple categorical determinations show near-perfect preservation.

We discover significant model-method interactions, with optimal pairings (e.g., Gemini + NeOn-CoT) achieving strong semantic retention while suboptimal combinations showed severe degradation. Our contributions are: (1) an evaluation framework for measuring semantic preservation in ontology learning, and (2) empirical evidence that semantic loss varies dramatically with model-method pairing, providing guidance for selecting optimal configurations in legal knowledge systems. These findings suggest that practitioners should empirically validate semantic preservation rather than assuming structural correctness guarantees downstream utility.

\printbibliography

\clearpage

\appendix

\setcounter{table}{0}
\renewcommand{\thetable}{A.\arabic{table}}
\renewcommand{\theHtable}{A.\arabic{table}}

\section{MAUD Tasks Reference}
\label{sec:maud_tasks}

This appendix provides the mapping between the task identifiers used throughout this paper and the corresponding questions from the MAUD task as implemented in LegalBench \cite{legalBench}. Table~\ref{tab:maud_tasks_complete} lists all 34 tasks with their full text.

\begin{table*}[!htbp]
\centering
\caption{MAUD Task Reference}
\label{tab:maud_tasks_complete}
\footnotesize
\begin{tabular}{lp{0.75\textwidth}}
\toprule
\textbf{Task ID} & \textbf{Question} \\
\midrule
t1 & Is the "ability to consummate" concept subject to Material Adverse Effect (MAE) carveouts? \\
t2 & How accurate must the fundamental representations and warranties be according to the bring down provision? \\
t3 & How accurate must the capitalization representations and warranties be according to the bring down provision? \\
t4 & When are representations and warranties required to be made according to the bring down provision? \\
t5 & How long is the additional matching rights period for modifications in case the board changes its recommendation? \\
t6 & What negative covenants does the requirement of Buyer consent apply to? \\
t7 & In case the Buyer's consent for the acquired company's ordinary business operations is required, are there any limitations on the Buyer's right to condition, withhold, or delay their consent? \\
t8 & Do changes in law that have disproportionate impact qualify for Material Adverse Effect (MAE)? \\
t9 & Do changes in GAAP or other accounting principles that have disproportionate impact qualify for Material Adverse Effect (MAE)? \\
t10 & Is Change of Recommendation permitted in response to an intervening event? \\
t11 & Is Change of Recommendation permitted as long as the board determines that such change is required to fulfill its fiduciary obligations? \\
t12 & What standard should the board follow when determining whether to change its recommendation in response to an intervening event? \\
t13 & What standard should the board follow when determining whether to change its recommendation in connection with a superior offer? \\
t14 & What is the knowledge requirement in the definition of "Intervening Event"? \\
t15 & What qualifies as a superior offer in terms of asset deals? \\
t16 & What qualifies as a superior offer in terms of stock deals? \\
t17 & Under what circumstances could the Board take actions on a different acquisition proposal notwithstanding the no-shop provision? \\
t18 & What type of offer could the Board take actions on notwithstanding the no-shop provision? \\
t19 & Is "financial point of view" the sole consideration when determining whether an offer is superior? \\
t20 & What is the Forward Looking Standard (FLS) with respect to Material Adverse Effect (MAE)? \\
t21 & Do changes caused by general economic and financial conditions that have disproportionate impact qualify for Material Adverse Effect (MAE)? \\
t22 & Does the wording of the Efforts Covenant clause include "consistent with past practice"? \\
t23 & How long is the initial matching rights period in case the board changes its recommendation? \\
t24 & How long is the initial matching rights period in connection with the Fiduciary Termination Right (FTR)? \\
t25 & Is an "Intervening Event" required to occur after signing? \\
t26 & What counts as Knowledge? \\
t27 & What is the liability standard for no-shop breach by target non-DO representatives? \\
t28 & What is the efforts standard? \\
t29 & Do pandemics or other public health events have to have disproportionate impact to qualify for Material Adverse Effect (MAE)? \\
t30 & Is there specific reference to pandemic-related governmental responses or measures in the clause that qualifies pandemics or other public health events for Material Adverse Effect (MAE)? \\
t31 & What carveouts pertaining to Material Adverse Effect (MAE) does the relational language apply to? \\
t32 & What is the wording of the Specific Performance clause regarding the parties' entitlement in the event of a contractual breach? \\
t33 & How long is the Tail Period? \\
t34 & What type of consideration is specified in this agreement? \\
\bottomrule
\end{tabular}
\vspace{0.3em}

\footnotesize
\textit{Note:} All MAUD tasks originate from LegalBanch \cite{legalBench}. Complete task definitions, including answer options and examples, are available in the \href{https://github.com/HazyResearch/legalbench}{LegalBench repository}.
\end{table*}

\end{document}